\documentclass{ecai}

\usepackage{graphicx}
\usepackage{latexsym}
\usepackage{amssymb}
\usepackage{array}
\usepackage[mathscr]{eucal}
\usepackage{multirow}
\usepackage{color}
\usepackage{bbm}

\usepackage[autostyle]{csquotes}
\usepackage{mathtools}
\usepackage{breqn}
\usepackage{amsmath}
\usepackage{booktabs}
\usepackage{algorithm}

\usepackage{soul}
\DeclarePairedDelimiterX{\norm}[1]{\lVert}{\rVert}{#1}

\usepackage[colorlinks=true,allcolors=black]{hyperref}
\usepackage{cleveref}
\usepackage[noend]{algpseudocode}
\usepackage{nicefrac}
\newtheorem{remark}{Remark}

\makeatother
\def\x{{\bf x}}

\def\y{{\bf y}}

\def\z{{\bf z}}
\def\W{{\bf W}}

\def\Y{{\bf Y}}
\def\Z{{\bf Z}}

\newtheorem{corollary}{Corollary}
\newtheorem{lemma}{Lemma}

\begin{document}
\begin{frontmatter}
\title{Piecewise-Stationary Combinatorial Semi-Bandit with Causally Related Rewards}
\author[]{\fnms{Behzad}~\snm{Nourani-Koliji}\orcid{0009-0001-2854-1955}}%
\author[]{\fnms{Steven}~\snm{Bilaj}}
\author[]{\fnms{Amir}~\snm{Rezaei Balef}}
\author[]{\fnms{Setareh}~\snm{Maghsudi}}
\address[]{University of Tübingen}

\begin{abstract}
We study the piecewise stationary combinatorial semi-bandit problem with causally related rewards. In our nonstationary environment, variations in the base arms' distributions, causal relationships between rewards, or both, change the reward generation process. In such an environment, an optimal decision-maker must follow both sources of change and adapt accordingly. The problem becomes aggravated in the combinatorial semi-bandit setting, where the decision-maker only observes the outcome of the selected bundle of arms. The core of our proposed policy is the Upper Confidence Bound (UCB) algorithm. We assume the agent relies on an adaptive approach to overcome the challenge. More specifically, it employs a change-point detector based on the Generalized Likelihood Ratio (GLR) test. Besides, we introduce the notion of \textit{group restart} as a new alternative restarting strategy in the decision making process in structured environments. Finally, our algorithm integrates a mechanism to trace the variations of the underlying graph structure, which captures the causal relationships between the rewards in the bandit setting. Theoretically, we establish a regret upper bound that reflects the effects of the number of structural- and distribution changes on the performance. The outcome of our numerical experiments in real-world scenarios exhibits applicability and superior performance of our proposal compared to the state-of-the-art benchmarks. 
\end{abstract}

\end{frontmatter}

\section{Introduction}
%
Multi-armed bandit (MAB) \cite{robbins1952some} is a class of sequential learning- and optimization problems. In the seminal MAB problem, the decision-maker (agent) selects one of the $K$ available arms, where each arm returns a reward drawn from a time-invariant, unknown distribution. The agent maximizes the total expected reward over the gambling horizon by using an effective decision-making strategy that maps the historical actions and outcomes to future actions. That is equivalent to minimizing the total expected \textit{regret}, which is the difference between the reward of the applied policy and that of the optimal policy in hindsight. Indeed, the MAB challenge boils down to the exploration-exploitation dilemma, where the agent decides between accumulating immediate rewards on the one side and obtaining information that might result in a larger reward only in the future on the other side. Due to its wide variety, the MAB framework is a potential candidate as a mathematical tool for tackling many real-world problems, for example, resource allocation in networks \cite{Maghsudi16:IWC}, recommender systems \cite{li2010contextual}, and clinical trials \cite{aziz2021multi}. 

The combinatorial multi-armed bandit (CMAB) problem is an extension of the seminal MAB. Instead of only one arm in each round, the agent chooses a number of them, i.e., it takes a combinatorial action. That results in exponential growth of the decision set by increasing the number of arms. Consequently, the conventional MAB methods such as UCB1 \cite{auer2002finite} become inefficient or inapplicable. In CMAB, we refer to each original arm as a {base arm}, and any subset of the base arms is a \textit{super arm}. Sometimes, the agent observes the reward of all base arms inside the super arm; In some other cases, the agent observes only one reward. The former type of feedback is a \textit{semi-bandit feedback}, whereas the latter is a \textit{bandit feedback}. The bandit problem becomes aggravated when a statistical structure influences the reward generation processes so that besides the excessively-large action set, the player deals with the structural relationships to decide optimally. We focus on combinatorial semi-bandit (CSB) problem with causally related rewards.

The seminal settings of CMAB- or CSB problems do not assume any statistical or probabilistic relationship between the base arms; Nevertheless, in several application domains, the potential dependency between the random variables can be abstracted by a structure. Despite being neglected for a long time, different types of the MAB problem with probabilistic or statistical relationships between the base arms, referred to as \textit{structured bandits} receive increasing attention from the research community in the past few years. For example, \cite{chen2016combinatorial}, \cite{wang2017improving}, and \cite{kveton2015cascading} assume that some arms may probabilistically be triggered based on the outcome of other arms. In \cite{lattimore2016causal}, prior knowledge about the causal structure that affects the rewards is available. 
The authors in \cite{nourani2022linear} introduce a causally structured CSB problem and use a directed acyclic graph to model the causal structure that influences the reward generation process. Their algorithm does not need apriori knowledge concerning the structural relationships as it can learn the structure from the streaming data. All of the works mentioned above study a stationary setting. 

Unlike the stationary stochastic setting, in many real-world scenarios, the reward distributions of base arms change over time in an evolving environment. For example, in recommender systems, the behavioral feedback of users is time-variant. It is possible to address the nonstationary behaviors of rapidly-varying environments using the adversarial bandit framework \cite{auer2002nonstochastic}. However, in some cases, the environment changes slowly and less frequently. In such scenarios, policies designed for stationary or adversarial bandits are sub-optimal. Generally speaking, there are two main approaches in modeling this type of nonstationarity in bandit problems; the \textit{switching case} (abruptly changing) \cite{zhou2020near} and the \textit{dynamic case} (smoothly changing) \cite{chen2021combinatorial} \cite{trovo2020sliding}. For the switching case, the reward distributions of base arms remain unchanged for certain intervals. The environment then varies if the distributions of a subset of base arms change instantly. The point where distributions change is a \textit{change-point} (or breakpoint) and an interval between any two consecutive change-point is a \textit{stationary segment} \cite{zhou2020near}. In contrast, in the dynamic case, the base arms' mean rewards evolve slowly instead of abruptly changing at one point, and the variation is bounded by a variation budget \cite{besbes2014stochastic}. In this paper, we focus on the switching case, also referred to as \textit{piecewise} stationary bandit model \cite{besson2019generalized}. We measure the decision-making performance using the notion of \textit{piecewise stationary regret}, i.e., the regret w.r.t. an oracle that knows the best action in each stationary segment.

In a piecewise stationary structured bandit problem, the reward generation processes might vary by changing the base arms' reward distributions and the structural relationships between the variables. Although such a model has remained unaddressed in the MAB literature, it accommodates several real-world applications. Those include financial markets, where not only the \textit{investors' stock purchasing behavior} but also the \textit{causal effects amongst the stock prices} can be time-varying \cite{shen2017tensor}. In such scenarios, an optimal investor follows both sources of change and adapts accordingly. While the availability of prior knowledge about the structural relationships is a strong and unrealistic assumption, inferring such structural relationships from the streaming partial feedback in the bandit setting is also challenging. We study a piecewise stationary structured CSB problem, where the causal relationships between the rewards and the distributions of the base arms evolve. In order to model the structural relationships, we rely on \textit{Structural Equation Models} \cite{giannakis2018topology}.

In general, there are two main approaches to follow the piecewise stationary behaviour of base arms distributions; \textit{passively adaptive} approach \cite{garivier2011upper} and \textit{actively adaptive} approach \cite{besson2019generalized} \cite{cao2019nearly} \cite{hartland2007change} \cite{liu2018change} \cite{zhou2020near} \cite{cheng2023distributed}. Methods of the former category are unaware of the change-points and rely on their understanding of the optimal action based on the most recent observations. On the contrary, methods of the latter category use a change detection algorithm to follow the distributions' changes and decide accordingly \cite{besson2019generalized}. Some studies show the superior performance of actively adaptive approaches \cite{mellor2013thompson}. Clearly, the performance of actively adaptive approaches rely significantly on the ability of the agent in handling the breakpoints. Current actively adaptive algorithms incorporate either \textit{global restart} or \textit{local restart} to restart learning the expected value of the instantaneous rewards of the base arms. The former method resets learning the expected values of all arms after detecting a change in one of them. The latter restarts learning only for those arms undergoing a change. These approaches suffer from a drawback as they ignore possible relationships amongst arms' distributions in making a decision upon restarting process. There are main reasons for introducing a new restarting strategy for bandit algorithms in structured piecewise stationary environments. Firstly, social networks, as one of the main target applications for bandit algorithms, exhibit large modularity measures \cite{newman2006modularity} \cite{borge2011structural}. Secondly, in some real-world scenarios changes within a network are not completely independent, but they are rather the result of the local spread of a change-seed within the network structure through mechanisms such as contagion \cite{easley2010networks}, social influence \cite{easley2010networks}, or diffusion \cite{thanou2017learning}, e.g. media-based marketing campaigns, or rumor diffusion over social networks \cite{shafipour2021identifying}. In this regard, we introduce the notion of \textit{group restart} where we restart the set of arms that are in the same group, upon detecting a change in any of them. We elaborate more on this in the following sections. We show the superior adaptation capabilities of this approach over local and global restarts in our experiments and discuss the effects of this approach over the upper bound of regret in theory. 

In this work, we introduce a piecewise stationary CSB problem with causally related rewards. Our framework accommodates the changes in the base arms' reward distributions and also in the causal relationships between the rewards. We provide an actively adaptive approach to tackle the problem. We introduce a novel alternative restarting strategy, namely \textit{group restart}, that can be used in the adaptation of stationary bandit algorithms to the piecewise stationary environments. We highlight the importance of using the knowledge of relationships amongst arms' distributions in our group restart strategy. We achieve this by showing its effects on the regret of the algorithm in dealing with the costly effects of both \textit{restarts} and \textit{delays of change point detectors}. Our algorithm uses a UCB-based policy for learning the expected rewards of the base arms and a GLR change-point detector. Furthermore, we integrate a mechanism in our algorithm to follow the changes of the causal graph structure that models the causal relationships between the rewards in the bandit setting. 
We provide the theoretical analysis of the regret upper bound for our algorithm. Our regret bound reflects the effects of both the number of causal graph changes and the number of distribution changes. Our numerical experiments using synthetic- and real-world data establish the advantage of our algorithm compared to the benchmarks. 

In \textbf{Section \ref{sec:prob-formulation}}, we introduce the piecewise stationary combinatorial semi-bandit problem with causally related rewards. In \textbf{Section \ref{sec:learning-algo}}, we develop our decision-making policy, namely, PS-SEM-UCB-Gr. \textbf{Section \ref{sec:theor-analysis}} presents the theoretical analysis of the regret performance of PS-SEM-UCB-Gr. \textbf{Section \ref{sec:exp-analysis}} includes the numerical experiments. \textbf{Section \ref{sec:conclusion}} concludes the paper with some suggestions for future works.

\section{Problem Formulation}\label{sec:prob-formulation}
In a \textbf{piece-wise stationary combinatorial semi-bandit (PSCSB)} problem with causally related rewards, a change from one stationary segment to the other results from varying (i) base arms' reward distributions or (ii) the causal relationships between rewards. The intervals with fixed reward distributions and static causal graph are distribution- and graph stationary segments, respectively. The change-points of both segment types appear randomly. 
We use $\mathcal{K}=\left\{1,\dots,\mathrm{K} \right\}$ to represent the set of $\mathrm{K}$ base arms, $\mathcal{D}\subseteq 2^{\mathcal{K}}$ the set of all super arms, and $\mathcal{T}=\left\{1,\dots,T \right\}$ a sequence of $T$ time-steps. Besides, $\theta_{k,t}$ is the distribution of the instantaneous reward of arm $k$ at time $t$ with mean ${\mu}_{k,t}$ and bounded support within $[0,1]$. The vector $\boldsymbol{\mu}_{t} = [\mu_{1,t}, \dots, \mu_{K,t}]$ is the expected values of the instantaneous rewards of all base arms at time $t$. Additionally, $\mathcal{A}_{t}$ is the underlying graph that shows the causal relations between the base arms' rewards. We use $\psi_{\boldsymbol{\mu}_{t},\mathcal{A}_{t}}(\x_{t})$ to denote the agent's expected reward from the decision vector $\x_{t}$ given  $\boldsymbol{\mu}_{t}$ and $\mathcal{A}_{t}$. Consequently, we characterize a PSCSB with the tuple $(\mathcal{K},\mathcal{D},\mathcal{T},\left\{\theta_{k,t} \right\}_{k\in \mathcal{K},t\in \mathcal{T}},\psi_{\boldsymbol{\mu}_{t},\mathcal{A}_{t}}(\x_{t}))$. Vector of base arms' instantaneous rewards at time $t$ is represented by $\mathbf{b}_{t} = [b_{1,t}, \dots, b_{K,t}] \in [0,1]^{K}$ and it follows a piece-wise independent and identically distributed (i.i.d.) model in each distribution stationary segment. A change to the distribution stationary segment of the environment corresponds to a change in at least one arm's reward distribution. Our setting assumes that the agent is given the meta information regarding the grouping (clustering) of arms such that arms within the same group tend to have their instantaneous rewards' distributions changed together. We use $g$ to denote a group of arms. $K_{g}=\left| g \right|$ is used to show the cardinality of the group $g$. $g_{k}$ represents the group to which arm $k$ belongs. We use $G$ to denote the set of all groups, $G = \left\{ g^{(1)},\dots,g^{(\zeta)} \right\}$, with $\left|G \right|=\zeta$ and $ \bigcup_{i\in [\zeta]}g^{(i)}=\mathcal{K}, \forall i,j \in [\zeta],g^{(i)} \cap g^{(j)} = \varnothing$ where $[\zeta]=\left\{1,\dots,\zeta \right\}$. $N_{\Theta}$ is used to denote the number of distribution stationary segments of the environment. We define the total number of distribution stationary segments for group $g$ as 
\begin{equation}
N_{g}=1+\sum_{t=1}^{T-1}\mathbbm{1}\left\{ \exists k\in g ~ \text{s.t.} ~ \theta_{k,t} \neq \theta_{k,t+1} \right\}.
\label{eq:group-change}
\end{equation}
Hence, the total number of stationary segments for all groups is $N_{G}=\sum_{g\in G}N_{g}$. This clarifies that $N_{G}$ can change depending on the way the grouping is performed. At each time $t$, the agent selects a \textit{decision vector} $\x_{t} = [x_{1,t}, \dots, x_{K,t}] \in \left \{ 0,1 \right \}^{K}$. We use $\mathcal{I}_{t}\subset \mathcal{K}$ to denote the set of chosen base arms $I_{t}\in \mathcal{K}$ at round $t$. We have $x_{k,t}=1$ if the base arm $k$ is in the super arm $\mathcal{I}_{t}$ at time $t$, otherwise $x_{k,t}=0$. The agent selects at most $m$ base arms at each time step. Hence, we define the set of all feasible decision vectors as $\mathcal{X} = \left \{ \x \mid \x \in \{0,1\}^{K} \wedge \left \| \x \right \|_{0}  \leq m \right\}$ where $\left\| \cdot \right \|_{0}$ determines the number of non-zero elements in a vector and the parameter $m$ is pre-determined. The causal relationships in the environment are modelled using a directed graph. More precisely, we consider an unknown piecewise static sparse Directed Acyclic Graph (DAG), $\mathcal{A}_{t} = (\mathcal{V}, \mathcal{E}_{t}, \W_{t})$.  $\mathcal{V}$ represents the set of $K$ vertices, i.e., $\left | \mathcal{V} \right | = K$, $\mathcal{E}_{t}$ and $\W_{t}$ denote the edge set and the weighted adjacency matrix at time $t$, respectively. We allow the edge set $\mathcal{E}_{t}$ to change arbitrarily every time the causal graph structure changes. However, the set of vertices $\mathcal{V}$ stays unchanged across time. This implies that the adjacency matrix $\W_{t}$ changes only in the elements $\W_{t}[i,j], \forall i,j \in \mathcal{K}, \forall t \in \mathcal{T}$, as far as the underlying graph structure remains a DAG without self-loop, i.e., $\W_{t}[i,i]=0, \forall i \in \mathcal{K}, \forall t \in \mathcal{T}$. $N_{\W}$ represents the number of graph stationary segments. An error-free piecewise static Structural Equation Model (SEM) \cite{giannakis2018topology} is used to model the generation of reward in the environment. At each time $t$, $\z_{t} = [z_{1,t}, \dots, z_{K,t}]$ is used to represent the exogenous input vector while $\y_{t} = [y_{1,t}, \dots, y_{K,t}]$ denotes the endogenous output vector of the SEM. We write,
\begin{align}
\label{eq:ExoInput}
    \z_{t} = \operatorname{diag}(\mathbf{b}_{t}) \x_{t},
\end{align}
where $\operatorname{diag}(\cdot)$ represents a diagonal matrix. This implies that the exogenous input $\z_{t}$ contains the semi-bandit feedback in the decision-making problem. We define the $k^{\text{th}}$ element of the endogenous output vector $\y_{t}$ at any time $t$ as
\begin{equation}
\label{eq:OverallRew}
y_{k,t} = \sum_{j=1}^{K} \W_{t}[k,j] y_{j,t} + z_{k,t},\quad\forall k\in \mathcal{K},
\end{equation}
At each time $t$, the endogenous output $y_{k,t}$ represents the \textit{overall reward} of base arm $k \in \mathcal{K}$. The element $\mathbf{W}[k,j]$ represents the causal effect of the overall reward of base arm $j$ on the overall reward of base arm $k$. Therefore, the overall rewards of base arms are causally related while the instantaneous reward of arm $k$ only directly contributes to the overall reward of arm $k$. It is important to distinguish between the relationships amongst arms' distributions and the causal relationships amongst the overall rewards. The first one only explains the prior information regarding the groupings of arms, while the second one is used in the mathematical formulation of the problem. 

The adjacency matrices $\W_{t}, \forall t\in \mathcal{T}$ are unknown a priori and $\W_{t}[i,j]\geq 0, \forall i,j \in \mathcal{K}, \forall t \in \mathcal{T}$. The matrix form of (\ref{eq:OverallRew}) at time $t$ is given as 
\begin{equation}
\label{eq:OverallRew-MatrixForm}
\y_{t} = \W_{t} \y_{t} + \z_{t}.
\end{equation}
As a result, we write $\y_{t}=(\mathbf{I}-\W_{t})^{-1}\operatorname{diag}(\mathbf{b}_{t}) \x_{t}$ by solving (\ref{eq:OverallRew-MatrixForm}) for $\y_{t}$, where $\mathbf{I}$ is the identity matrix. We assume that the agent is able to observe both the instantaneous semi-bandit feedback vector $\z_{t}$ and the overall reward feedback vector $\y_{t}$. The \textit{payoff} received by the agent upon choosing the decision vector $\x_{t}$ is defined as
\begin{equation}\label{eq:reward}
r_{t}(\x_{t}) = {\bf c}^{\top}\y_{t} = {\bf c}^{\top} (\mathbf{I}-\W_{t})^{-1} \operatorname{diag}(\mathbf{b}_{t}) \x_{t}, 
\end{equation}
where ${\bf c}=[c_{1}, \dots, c_{K}] \in \left \{ 0,1 \right \}^{K}$ is pre-determined. The agent is interested in the output $y_{k}$ in the causal network if $c_{k}=1$, and $c_{k}=0$ otherwise. Since the graph $\mathcal{A}_{t}$ is a DAG, the adjacency matrix $\W_{t}$ is nilpotent. This property guarantees that the matrix $(\mathbf{I}-\W_{t})$ is invertible. 
Given a decision vector $\x_{t} \in \mathcal{X}$, the expected payoff at time $t$ is calculated as
\begin{equation}
\label{eq:ExpPayoff}
\psi_{\boldsymbol{\mu}_{t},\mathcal{A}_{t}}(\x_{t}) = \mathbb{E}\left [  r_{t}(\mathbf{X}) | \mathbf{X} = \x_{t} \right ],
\end{equation}
where the expectation concerns the randomness in the reward generating process. We denote by $\x_{t}^{*}=\underset{\mathbf{x} \in \mathcal{X}}{\text{argmax}} ~\psi_{\boldsymbol{\mu}_{t},\mathcal{A}_{t}}(\mathbf{x})$ the decision vector with maximum expected reward at time $t$. The agent minimizes the cumulative piecewise stationary regret defined as
\begin{equation}
\mathcal{R}(T)=\mathbb{E}\left [ \sum_{t=1}^{T}(\psi_{\boldsymbol{\mu}_{t},\mathcal{A}_{t}}(\x_{t}^{*})- \psi_{\boldsymbol{\mu}_{t},\mathcal{A}_{t}}(\x_{t})) \right ].\label{eq:Expregret}
\end{equation}
%
\section{The Learning Algorithm}
\label{sec:learning-algo}
In this section, we develop a solution to the formulated problem. We first introduce the group restart strategy, and the online graph learning. Afterward, we present our decision-making policy, namely, PS-SEM-UCB-Gr.  
\subsection{Group Restart Strategy.}
\label{subsec:GroupRestart}
Restarting process plays a key role in the decision making strategy in piecewise stationary bandit algorithms. Upon taking the global restart strategy, the agent's regret increases due to the costly effects of restarting of all arms. Moreover, by taking local restart strategy, delays of change point detectors for different arms can make the algorithm to incur linear regret in some intervals. One way to address these issues is in the structured environments where changes are not always completely independent and having side information w.r.t. relationships between arms' distributions can be helpful in making decisions upon restarts. 
There are certain research directions in MAB literature where relationships amongst the arms are considered. For instance, in \cite{valko2014spectral}, it is assumed that each item that the algorithm recommends is a node of a known graph and the expected rating of the neighboring nodes are similar. Furthermore, in \cite{gentile2014online}, it is suggested that the nodes of the graph can be clustered according to some apriori unknown clustering and the arms within the same cluster exhibit similar behaviours. Also, in \cite{yang2020laplacian}, the relationship between the users is captured by an underlying graph and user preferences are assumed to have smooth signals on the graph. In such settings, it is natural to anticipate that if an arm's expected reward is changed, then due to the relationships of the arms, the set of arms that are closely connected to it go through changes as well. Consequently, we propose \textit{group restart} strategy as an efficient alternative in structured environments where grouping information might be either available in advance or learned from the data \cite{gentile2014online} \cite{li2016collaborative}. As the result of our theoretical analysis, we show that a structure-based grouping in group restart strategy can help to reduce the regret upper bound compared to local and global restarts. 
\subsection{Piece-wise Static Graph Learning.}
\label{subsec:GraphLearning}
Considering the required knowledge of $\W_{t}$ in finding the optimal decision vector, we propose an online graph learning framework that uses the collected feedback $\y_{t}$ and $\z_{t}$ and allows for modelling both the random and the smooth transitions of the causal graph. At each time $t$, we stack the feedback, from the last graph-change-point up to the current time, as consecutive columns in $\Z_{t}$ and $\Y_{t}$, hence, $\Y_{t} = \W_{t} \Y_{t} + \Z_{t}$. We use the collected feedback history, $\Y_{t}$ and $\Z_{t}$, as the input to a parametric graph learning algorithm for a static SEM \cite{giannakis2018topology}. Formally, the adjacency matrix at time $t$ is the solution to the following optimization problem: 
\begin{equation}
\label{eq:Optimization-A}
\begin{aligned}
\hat{\W}_{t} = \underset{\W \in \mathbb{R}^{K\times K}}{\text{argmin}} \quad & \norm{\Y_{t} - \W \Y_{t} - \Z_{t}}_{\text{F}}^{2} +\lambda_{1} \left \| \W \right \|_{1}\\
\textrm{s.t.} \quad  &\W[k,k]=0, \forall k \in \mathcal{K} \\
\end{aligned}
\end{equation}
where $\left \| \cdot \right \|_{\text{F}}$ represents the Frobenius norm of matrices. The symbol $\left \| \cdot \right \|_{1}$ denotes the $L^{1}\text{-norm}$ of the matrices and it is used to impose sparsity over the estimated adjacency matrix $\hat{\W}_{t}$. We use the notation $\hat{\W}^{(i)}$ to represent the estimated adjacency matrix for the $i^{\textit{th}}$ static graph. In order to impose slow topological variations across time, from one static graph to the next, one may add a second regularization term $\lambda_{2} \left \| \hat{\W}^{(i+1)} - \hat{\W}^{(i)} \right \|_{1}$ in (\ref{eq:Optimization-A}) and have a form of the optimization problem in (\ref{eq:Optimization-A}) that stays convex. This second regularization allows the algorithm to penalize deviation of the current graph estimate from the predecessor, hence implementing a transfer of knowledge that is gained from the previous segment. 
\subsection{The PS-SEM-UCB-Gr Algorithm}
\begin{algorithm}[t]
    \caption{Graph Learning Data Generation (GLDG)}
    \begin{algorithmic}[1]
        \State Create an initialization matrix $\textbf{init}$, $\Y_{0}=[]$, $\Z_{0}=[]$.
        \For{$t'=1,2,\dots,K$}
        \State $\x_{t}:=\textbf{init}[:,t']$
        \State Play $\mathcal{I}_{t}$, receive reward $r(\x_{t})$, ${s}_{I_{t},n_{I_{t},t}} \leftarrow z_{I_{t},t}$,$\forall I_{t} \in \mathcal{I}_{t}$.
        \For{$\textbf{all} ~ I_{t}\in \mathcal{I}_{t}$}
        \State update: $\hat{{\mu}}_{I_{t},t}$ using (\ref{eq:AvgRew}), ${n}_{I_{t},t}$ using (\ref{eq:Counter-BaseArm}).
        \If {$\textbf{GLR}(s_{I_{t},1},\dots,s_{I_{t},n_{I_{t},t}};\delta)=1$}  
        \State $\forall k \in {g}_{I_{t}}$: ${n}_{k,t}\leftarrow 0$, $\hat{{\mu}}_{k,t}\leftarrow 0$, ${\tau}_{k} \leftarrow t$.
        \State $\tau' \leftarrow t $, $\Omega \leftarrow \Omega \cup {g}_{I_{t}}$. 
        \EndIf
        \EndFor
        \For{$\textbf{all} ~ k\in \mathcal{K}$}
        \If{${n}_{k,t}\neq 0$} 
        \State ${U}_{k,t}\leftarrow  \hat{{\mu}}_{k,t}+\sqrt{\frac{(m+1)\log(t-{\tau}_{k})}{n_{k,t}}}$  
        \EndIf
        \EndFor
        \State $[\Y_{t}]\leftarrow[\Y_{t-1} ,\y_{t}]$, $[\Z_{t}]\leftarrow[\Z_{t-1} ,\z_{t}]$, $t \leftarrow t+1$
        \EndFor
        \State Solve (\ref{eq:Optimization-A}) to get $\hat{\W}_{t-1}$.
        \State $\textit{flag}=0$
    \end{algorithmic}
    \label{algo:GLDG}
\end{algorithm}
In this section, we describe our decision-making policy. Its core is the Upper Confidence Bound policy. Besides, we use two previously-proposed methods, namely \textit{group restart}, and \textit{piece-wise static graph learning}. Finally, we integrate a mechanism for detecting the changes to the adjacency matrix of the causal graph. Each time the algorithm decides to infer the new adjacency matrix, it starts a subroutine inside the main algorithm to obtain $K$ data samples by interacting with the new environment. It is crucial that the new dataset satisfies the conditions for the precise inference and unique identification of the new graph adjacency matrix \cite{bazerque2013identifiability} \cite{nourani2022linear}. We refer to this subroutine as \textit{Graph Learning Data Generation} (GLDG). For these $K$ rounds, PS-SEM-UCB-Gr picks $K$ columns of an \textit{initialization matrix}, namely, $\mathbf{Init}\in \{0, 1\}^{K \times K}$ in a sequential way where $\mathbf{Init}$ is created as described in \cite{nourani2022linear}, Section $3.2$. Based on the discussion above, we assume that there are at least $K+1$ rounds between any two consecutive changes in the graph. That guarantees sufficient time to infer the new ground truth graph after every change. We refer to the rounds inside a GLDG phase as \textit{graph initialization} rounds and the rest as \textit{normal} rounds. In every round, the GLR change-point detectors and the UCB index developments are working. 
\begin{algorithm}[t]
\begin{algorithmic}[1]
\State \textbf{Initialization:} $\forall k\in \mathcal{K}$, $n_{k,0} \leftarrow 0$, $\hat{{\mu}}_{k,0}\leftarrow 0$, ${\tau}_{k}=0$, $t=1$, $\tau'=0$, $\textit{flag}=1$. Get $G=\left\{{g}^{(1)},\dots,{g}^{(\zeta)} \right\}$
\While{$t\leq T$} 
\If{$\textit{flag}=1$} 
\State Run \textit{\textbf{GLDG}.} 
\EndIf
\If{$\Omega \neq \varnothing$ }
\State Pick $a\in \Omega$, Randomly choose $\mathcal{I}_{t}$ with $a\in \mathcal{I}_{t}$. 
\State Remove $a$ from $\Omega$.
\Else 
\State Solve (\ref{eq:Optimization-x}) for $\x_{t}$.
\EndIf
\State Play $\mathcal{I}_{t}$, receive reward $r(\x_{t})$, ${s}_{I_{t},n_{I_{t},t}} \leftarrow z_{I_{t},t}$,$\forall I_{t} \in \mathcal{I}_{t}$.
\For {$\textbf{all} ~ I_{t}\in \mathcal{I}_{t}$}
\State update: $\hat{{\mu}}_{I_{t},t}$ using (\ref{eq:AvgRew}), ${n}_{I_{t},t}$ using (\ref{eq:Counter-BaseArm}).
\If{$\textbf{GLR}(s_{I_{t},1},\dots,s_{I_{t},n_{I_{t},t}};\delta)=1$} 
\State $\forall k \in {g}_{I_{t}}$: ${n}_{k,t}\leftarrow 0$, $\hat{{\mu}}_{k,t}\leftarrow 0$, ${\tau}_{k} \leftarrow t$. 
\State $\tau' \leftarrow t$, $\Omega \leftarrow \Omega \cup {g}_{I_{t}}$.
\EndIf
\EndFor
\If{$\exists ~ c\in \mathbb{N}: t-\tau' = c \left \lfloor \frac{K}{p} \right \rfloor$}
\State  $\Omega=\bigcup_{i\in [\zeta]}{g}^{(i)}$
\EndIf
\For {\textbf{all} ~ $k\in \mathcal{K}$}
\If{${n}_{k,t}\neq 0$}
\State ${U}_{k,t}\leftarrow  \hat{{\mu}}_{k,t}+\sqrt{\frac{(m+1)\log(t-{\tau}_{k})}{{n}_{k,t}}}$ 
\EndIf
\EndFor
\State $[\Y_{t}]\leftarrow[\Y_{t-1} ,\y_{t}]$, $[\Z_{t}]\leftarrow[\Z_{t-1} ,\z_{t}]$
\If{$ \left\| \y_{t}-\hat{\W}_{t-1}\y_{t}-\z_{t}\right\|_{2}^{2} > \epsilon$}  
\State $\textit{flag}=1$, $\Y_{t}=[]$, $\Z_{t}=[]$
\Else
\State Solve (\ref{eq:Optimization-A}) to get $\hat{\W}_{t}$.
\EndIf
\State $t \leftarrow t +1$
\EndWhile
\end{algorithmic}
\caption{PS-SEM-UCB\textcolor{black}{-Gr}: Piecewise Stationary - Structural Equation Model - Upper Confidence Bound - Group Restart}
\label{algo:ps-sem-ucb}
\end{algorithm}
The input parameters of PS-SEM-UCB-Gr include the number of steps ($T$), number of arms ($K$), uniform exploration probability $p\in (0,1)$, and $\delta$ as the confidence level of the GLR change-point detector. The policy uses the parameter $\tau'$ to perform the uniform forced exploration over all base arms in Line $16$ of Algorithm \ref{algo:ps-sem-ucb}. The forced uniform exploration guarantees that the GLR change-point detectors receive sufficient samples. Considering that we are using group restart, UCB developments of arms from different groups might have different resetting times. Therefore, the policy uses $\boldsymbol{\tau}= [{\tau}_{1}, \dots,{\tau}_{K}] $ to manage the restarting times of UCB indices. The variable \textit{flag} is used to call the GLDG subroutine. For any arm $k$, the empirical average of the instantaneous rewards at any time $t=t_{1}$ w.r.t. its last restarting time at $t={\tau}_{k}$ yields
\begin{equation}
\label{eq:AvgRew}
\hat{{\mu}}_{k,t_{1}} = \frac{\sum_{t=\tau_{k}+1}^{t_{1}} {z}_{k,t} }{{n}_{k,t_{1}}},
\end{equation}
where ${n}_{k,t_{1}}$ is the number of times that the base arm $k$ is observed up to time $t=t_{1}$ since its last restart at $t={\tau}_{k}$. Formally,
\begin{equation}
\label{eq:Counter-BaseArm}
{n}_{k,t_{1}} = \sum_{t=\tau_{k}+1}^{t_{1}} {x}_{k,t}.
\end{equation}
The set $\Omega$ holds the index of those arms whose UCB developments are being restarted or that are candidates for forced exploration. After the graph initialization period, in each round, PS-SEM-UCB-Gr first checks the set $\Omega$, in Line $5$, otherwise the agent plays the next super arm according to the result of the combinatorial optimization in Line $9$. The combinatorial optimization uses the current UCB indices and the last estimate of the causal graph. We denote the UCB index of base arm $k$ at time $t$ as ${U}_{k,t}$ such that we have the UCB indices of all base arms in the vector $\mathbf{U}_{t} = [{U}_{1,t}, \dots,{U}_{K,t}]$. Therefore, the combinatorial optimization for finding the best decision vector yields
\begin{equation}
\label{eq:Optimization-x}
\begin{aligned}
\mathbf{x}_{t} = \underset{\mathbf{x} \in \mathcal{X}}{\text{argmax}} \quad & \mathbf{c}^{\top} (\mathbf{I} - \hat{\W}_{t-1})^{-1} \operatorname{diag}(\mathbf{U}_{t-1}) \mathbf{x}.
\end{aligned}
\end{equation}
Let $\mathbf{M}^{\top} = \mathbf{c}^{\top} (\mathbf{I} - \hat{\W}_{t-1})^{-1} \operatorname{diag}(\mathbf{U}_{t-1})$. The elements of $\hat{\W}_{t-1}$, $\mathbf{c}$, and $\mathbf{U}_{t-1}$ are non-negative, then the optimization problem (\ref{eq:Optimization-x}) can be solved by finding a subset of elements in $\mathbf{M}$ such that $\mathbf{x} \in \mathcal{X}$. Therefore, it is solvable by using an efficient sorting algorithm that ranks the elements of $\mathbf{M}$. Consequently, the agent plays $\mathbf{x}_{t}$, collects the reward in Line $10$ according to (\ref{eq:reward}), and updates the vectors $\hat{\boldsymbol{\mu}}_{t}$ and $\mathbf{n}_{t}$ in Line $12$. The notation ${s}_{I_{t},n_{I_{t},t}} \leftarrow z_{I_{t},t}$ in Line $10$ implies that the collected feedback $z_{I_{t},t},\forall I_{t}\in \mathcal{I}_{t}$ is the sample number $n_{I_{t},t}$ in the sequence of samples for arm $I_{t}$ since its last restart at $t={\tau}_{I_{t}}$. We use the \textbf{GLR} change-point detector \cite{besson2019generalized} defined as 
\begin{multline}
    \textbf{GLR}(s_{1},\dots,s_{n};\delta) \coloneqq \\
    \mathbbm{1} \{ \text{sup}_{\alpha\in [1,n-1]}[\alpha \times \text{kl}(\hat{\beta}_{1:\alpha},\hat{\beta}_{1:n})+\\
    \left(n-\alpha\right)\times \text{kl}(\hat{\beta}_{\alpha+1:n},\hat{\beta}_{1:n})]\geq \gamma(n,\delta)\},
\end{multline}
where $\hat{\beta}_{\alpha:\alpha^{'}}$ is the mean of the observations between $\alpha$ and $\alpha^{'}$, $\text{kl}(x,y)=x\log\left(\frac{x}{y}\right)+(1-x)\log\left(\frac{1-x}{1-y}\right)$ is the binary relative entropy between any two Bernoulli distributions with means $x$ and $y$. The function $\gamma(n,\delta)$ is the threshold function for the GLR test. Theoretically, we choose this threshold function following Lemma $2$ in \cite{besson2019generalized}. However, in all our numerical experiments, we follow \enquote{Practical considerations} in \cite{besson2019generalized}, and select $\gamma(n,\delta)=\text{ln}(\frac{3n\sqrt{n}}{\delta})$. If $\textbf{GLR}(s_{1},\dots,s_{n};\delta)=1$, the algorithm applies group restarts in Line $14$. The algorithm updates the UCB indices in Line $20$. In Line $22$, the graph-change detection mechanism uses two vectors of $\mathbf{z}_{t}$ and $\mathbf{y}_{t}$ to test the validity of the last estimate of the graph adjacency matrix, $\hat{\W}_{t-1}$. If the error value for the graph-change detection formulation exceeds $\epsilon$, then the algorithm notifies that the previous estimate of the graph structure is no longer valid. Consequently, in Line $23$, we have $\text{flag}=1$, and the previously collected sets of feedback in $\mathbf{Z}_{t}$ and $\mathbf{Y}_{t}$ are dropped. Parameter $\epsilon$ represents the error we accept in the graph estimation process. In this paper, we take $\epsilon=0$. However, assuming $\epsilon\neq 0$, the effects of $\epsilon$ should be considered in the regret analysis. In case the collected feedback vectors $\mathbf{y}_{t}$ and $\mathbf{z}_{t}$ satisfy the SEM formulation for $\hat{\W}_{t-1}$, in Line $21$, PS-SEM-UCB-Gr  uses the newly updated matrices $\mathbf{Y}_{t}$ and $\mathbf{Z}_{t}$, to improve the adjacency matrix estimation. It is important to notice that the algorithm does not restart the UCB development upon detecting a graph-change. It also does not restart the graph learning following any distribution-change detection. 
\section{Theoretical Analysis}
\label{sec:theor-analysis}
In this section, we deliver the analysis for the expected regret of PS-SEM-UCB-Gr algorithm. We perform the regret analysis according to any grouping of arms, with local and global restarts as special cases. We denote the maximum delay across all detected changes as $d$. We divide the time line into stationary segments of base arm distributions. The graph changes will be treated separately as they do not affect the UCB developments and only contribute to the regret in terms of a constant, based on the graph-learning phase. We also define the suboptimality gaps in our setting as the reward difference between the optimal decision vector $\x^*$ and an arbitrary decision vector $\x$: $\Delta_t(\x)=\psi_t(\x_t^*)-\psi_t(\x)$, where $\psi_t(\x)$ is the mean reward of $\x$, with only subscript $t$ used to parameterize it for better readability. The largest gap is denoted as $\Delta_{\max}=\max_t\max_{\x:\psi_t(\x)<\psi_t(\x_t^*)}\Delta_t(\x)$, and the smallest $\Delta_{\min}=\min_t\min_{\x:\psi_t(\x)<\psi_t(\x_t^*)}\Delta_t(\x)$. As it is essential for estimating the total regret bound, we deliver the regret for the stationary case, with an improvement over the work of \cite{nourani2022linear}, as our result does not scale with the number of layers in the causal graph but only with the total number of arms;

\begin{lemma}
Let $\omega_t^T=\mathbf{c}^{\top} (\mathbf{I} - \hat{\mathbf{W}}_{t-1})^{-1} \operatorname{diag}(\x_{t+1})$ and $\omega_{\max}=\max_t\max_k \omega_{k,t},k\in \mathcal{K}$. In the stationary case $(N_{\Theta}=1 \wedge N_{\W}=1)$ of the PS-SEM-UCB-Gr algorithm, the upper regret bound is given as: \begin{equation*}
    \mathcal{R}(T)\leq \left[\frac{4\omega^2_{\max}m^2(m+1)K \log(T)}{\Delta^2_{\min}}+\frac{\pi^2}{3}m K + K\right]\Delta_{\max},
\end{equation*}
with $\Delta_{\max}$ as the largest suboptimality gap and $\Delta_{\min}$ smallest suboptimality gap.
\label{lemma:stationary_case}
\end{lemma}

The adapted proof is given in the supplementary materials. The following \Cref{theorem-nonstationary} states a bound on the regret in the non-stationary case of our proposed decision-making policy for any grouping of arms. 

\begin{theorem}
Let $\omega_t^T=\mathbf{c}^{\top} (\mathbf{I} - \hat{\mathbf{W}}_{t-1})^{-1} \operatorname{diag}(\x_{t+1})$ and $\omega_{\max}=\max_t\max_k \omega_{k,t},k\in \mathcal{K}$. The expected regret of the PS-SEM-UCB-Gr policy is upper bounded as: 
\begin{align*}
\mathcal{R}(T)&\leq \sum_{g\in G}\left[N_{g} K_{g} R_0({T})+(\delta T+1+\frac{\pi^2m}{3}) N_{g} K_g\Delta_{\max}\right]\\
    &+\left(Tp+dN_{G}+\delta T (K+N_{G})+N_{\W} K \right)\Delta_{\max},
\end{align*}
with $R_0(T)=\frac{4\omega^2_{\max}m^2(m+1) \log(T)}{\Delta^2_{\min}}\Delta_{\max}$.
\label{theorem-nonstationary}
\end{theorem}
\paragraph{}
See Section 1 of supplementary material for the proof.

This is the general regret upper bound that reflects the importance of grouping of arms. We are able to retrieve the bounds according to the given grouping of arms. We assumed the knowledge of groupings of base arms based on structural relationships between arms' distributions.

Following local restart strategy, $G=G_{\text{local}}$, we have $K_g=1$, $\forall g \in G_{\text{local}}$ and $|G_{\text{local}}|=K$, thus $\sum_g N_g K_g=N_{G_{\text{local}}}$. If we follow global restart strategy, $G=G_{\text{global}}$, we have $K_g=K$, $\forall g\in G_{\text{global}}$ and $|G_{\text{global}}|=1$, thus $\sum_g N_g K_g = KN_{G_{\text{global}}}$. It is important to note that the number of restarts differs for local and global restart strategies, since $N_{G_{\text{global}}}\leq N_{G_{\text{local}}}$. In the following, we compare the performance of our approach with local restarts and global restarts on the amount of regret increase in the distribution stationary segment after a breakpoint.  
\begin{remark}\label{Ramark}
We rewrite the regret upper bound in \Cref{theorem-nonstationary} as $R(T) \leq \Sigma_{g \in G} [C_1 N_g K_g+C_2 N_{\mathrm{g}}]+C_3$ where $C_1, C_2, C_3$ are independent of the grouping of arms. Let us assume that the breakpoint $\nu$ happens from $t$ to $t+1$ with change to $\mathfrak{K}_{\nu}$ arm distributions that belong to $\eta_{\nu}$ groups (clusters). The increase of the regret value within the stationary segment after breakpoint $\nu$ can be written as $\Delta R(\nu) \leq C_1 \Sigma_{g \in G} K_g \mathbbm{1}\left\{ \exists k\in g ~ \text{s.t.} ~  \theta_{k,t} \neq \theta_{k,t+1} \right\} + C_2 \Sigma_{g \in G} \mathbbm{1}\left\{ \exists k\in g ~ \text{s.t.} ~ \theta_{k,t} \neq \theta_{k,t+1} \right\} $. Consequently, we have the followings;
\begin{itemize}
\item In the case of \textit{Local restart}, we have $\Delta R(\nu) \leq C_{1}\mathfrak{K}_{\nu}+C_{2}\mathfrak{K}_{\nu}$.
\item In the case of \textit{Global restart}, we have $\Delta R(\nu) \leq C_{1}K+C_2$.
\item If the total number of arms inside the $\eta_{\nu}$ groups is $\mathfrak{K}_{\nu}$, for \textit{Group restart}, we have $\Delta R(\nu) \leq C_{1}\mathfrak{K}_{\nu}+C_{2}\eta_{\nu}$. 
\item If in the $\eta_{\nu}$ groups, there are collectively $s$ arms whose distributions did not change at $\nu$, in this case, for \textit{Group restart} we have $\Delta R(\nu) \leq C_{1}(\mathfrak{K}_{\nu}+s)+C_{2}\eta_{\nu}$.
\end{itemize}
\end{remark}
In the above, the first term, scaling with $C_{1}$, is the regret due to number of restarted arms, while the second term, scaling with $C_{2}$, is affected by the delays. These results clarify the idea behind using a group restart strategy, especially in cases where the $\mathfrak{K}_{\nu}$ changed arms are from a small number of $\eta_{\nu}$ clusters. Intuitively, in networks with high modularity measures, we can expect to have smaller number for $s$ and a better performance for the group restart strategy. 

By the following corollary, through fine-tuning the hyper-parameters $\delta$ and $p$ and with the assumption of the prior knowledge of $N_{G}$, we can achieve a sub-linear regret bound; 
\begin{corollary}
Let $\Delta_{\min}^{\text{change}}=\min_{i}\max_{k\in\mathcal{K}}|\mu_{k,i}-\mu_{k,i-1}|$.
By choosing $\delta=\frac{1}{T}$ and $p=\sqrt{\frac{N_G K\log{T}}{T}}$, the regret is upper-bounded by the following,  
\begin{align*}
    \mathcal{O}\left(\left(\frac{\sum_{g\in G}N_gK_g\log{T}}{\Delta_{\min}}+\frac{\sqrt{N_G KT\log{T}}}{\left(\Delta_{\min}^{\text{change}}\right)^2}+N_{\mathbf{W}}K\right)\Delta_{\mathrm{max}}\right)
\end{align*}
\end{corollary}
Our regret bound shows an improvement in comparison to the result of \cite{zhou2020near} in terms of the dependency of total restarts $N_G$, even though our algorithm does not require the prior knowledge of the causal graphs. In the absence of graph-changes, the respective contribution to the regret stems solely from the very first initialization, i.e., $N_{\W}=1$.
\section{Experimental Analysis}
\label{sec:exp-analysis}
In this section, we evaluate the performance of our proposed decision-making policy using synthetic- and real-world datasets by comparing it with the following state-of-the-art combinatorial semi-bandit algorithms as benchmarks; \textbf{CTS} \cite{huyuk2019analysis} is a Thompson sampling-based algorithm for stationary environments; \textbf{GLR-CUCB} \cite{zhou2020near} is a UCB-based algorithm for piecewise stationary environments. It employs a GLR change-point detector and uses a global restart strategy. We implemented the same algorithm with local restarts and group restarts, GLR-CUCB-Lo and GLR-CUCB-Gr, respectively; \textbf{CUCB-SW} \cite{chen2021combinatorial} is an algorithm that uses a sliding window to follow the base arms' distribution changes while developing the corresponding UCB indices; \textbf{Orc-R} is PS-SEM-UCB-Gr with the Oracle-Restart. This algorithm is given the prior information w.r.t. all distribution change-points and it only restarts the groups where a change is detected. In addition, we implement the PS-SEM-UCB-Gl and PS-SEM-UCB-Lo that are working based on global restart and local restart strategy, respectively. 

All three algorithms \textit{CTS}, \textit{GLR-CUCB}, and \textit{CUCB-SW} require access to the exact- or to an approximation oracle that solves the combinatorial optimization (\ref{eq:Optimization-x}); that is, they need prior knowledge of the ground truth causal graph at any time. Such a strong assumption renders them inapplicable in the absence of such prior knowledge. For a fair comparison, we apply all benchmarks to the instantaneous rewards feedback vector $\z_{t}$ at each time $t$. We implemented the exact optimization oracles for \textit{CTS}, \textit{GLR-CUCB}, \textit{GLR-CUCB-Gr}, \textit{GLR-CUCB-Lo}, and \textit{CUCB-SW}.  
\begin{figure}[t]
    \centerline{
    \includegraphics*[width=0.53\textwidth]{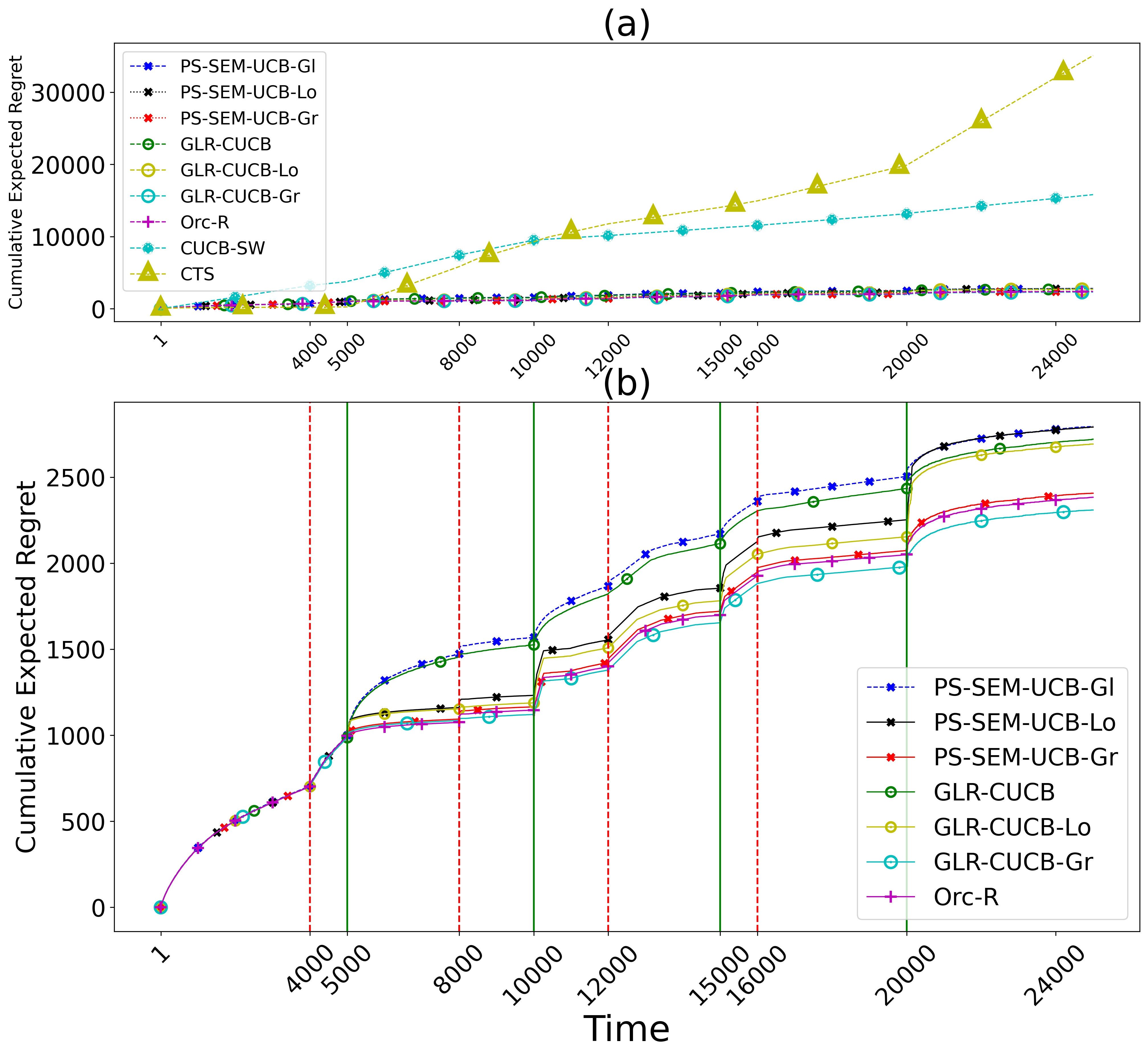}
    }
    \caption{Cumulative Expected Regret.}
    \label{fig:cum-reg}
\end{figure}
%
\subsection{Synthetic Dataset}
In the following, we describe the synthetic dataset used in the experiments. It has $4$ graph-change-points and $4$ distribution-change-points. For all different graph structures, we have $K = 18$ nodes. We draw the elements of the adjacency matrices $\mathbf{W}_{t}$ from a uniform distribution over $[0.1,0.9]$. The edge density of the ground truth adjacency matrices is $0.15$. The $K = 18$ arms are divided into $3$ groups of $6$ arms. We select $m=4$ in this experiment and $T=25000$. At each time $t$, the vector of instantaneous rewards $\mathbf{b}_{t}$ follows a multivariate normal distribution with the support in $[0,1]^{18}$ and a spherical covariance matrix. In supplementary material, Figure $1$ visualizes the expected values of base arms' rewards across time, and Figure $2$ presents the visualization of optimal super arm across time. As shown in Section \ref{sec:prob-formulation}, the reward generation process follows the SEM in (\ref{eq:OverallRew}). All distribution-stationary-segments of the environment have the same lengths. The regularization parameter $\lambda_{1}$ is tuned by grid search over $[0.0001,10000]$. We evaluate the estimated adjacency matrix at each time $t$ by using the mean squared error defined as $\text{MSE} = \frac{1}{K^{2}} \left\| \mathbf{W}_{t} - \hat{\mathbf{W}}_{t} \right\|_{\text{F}}^{2}$. \Cref{fig:cum-reg}-a shows the poor performance of \textit{CUCB-SW}, and \textit{CTS}, compared to other algorithms. In \Cref{fig:cum-reg}-b, we highlight the differences in the performance of \textit{Orc-R}, PS-SEM-UCB, \textit{GLR-CUCB} under various restarting strategies. One can observe the better performance of PS-SEM-UCB-Gr compared to \textit{GLR-CUCB}. That happens although PS-SEM-UCB-Gr does not require prior knowledge of the distribution-change-points and the causal graphs. The effects of different restarting strategies can be observed as well. Global restarts adds to the regret significantly by restarting the entire set of base arms. On the opposite, local restart suffers from the delay on those breakpoints where the number of changed distributions is large. In \Cref{fig:cum-reg}-b, each vertical green solid line represents the time of a distribution-change, and each vertical red dashed line represents a graph-change. 
\subsection{Real-World Application}
\begin{figure}[t]
\centerline{
\includegraphics*[width=0.49\textwidth]{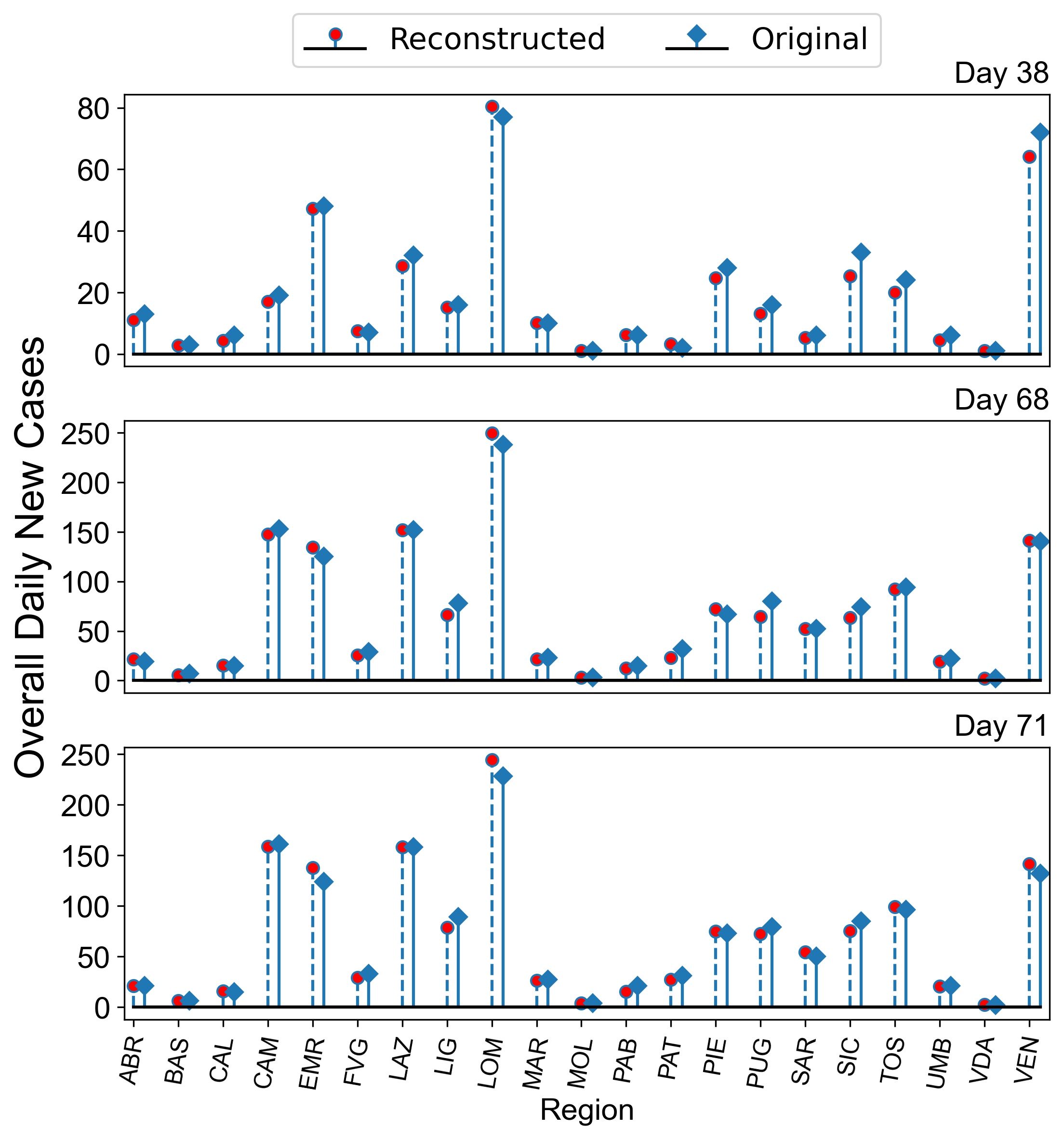}}
\caption{Original and reconstructed daily new cases.}
\label{fig:sig-reconstruct}
\end{figure}

\begin{figure}[t]
\centerline{
\includegraphics*[width=0.55\textwidth]{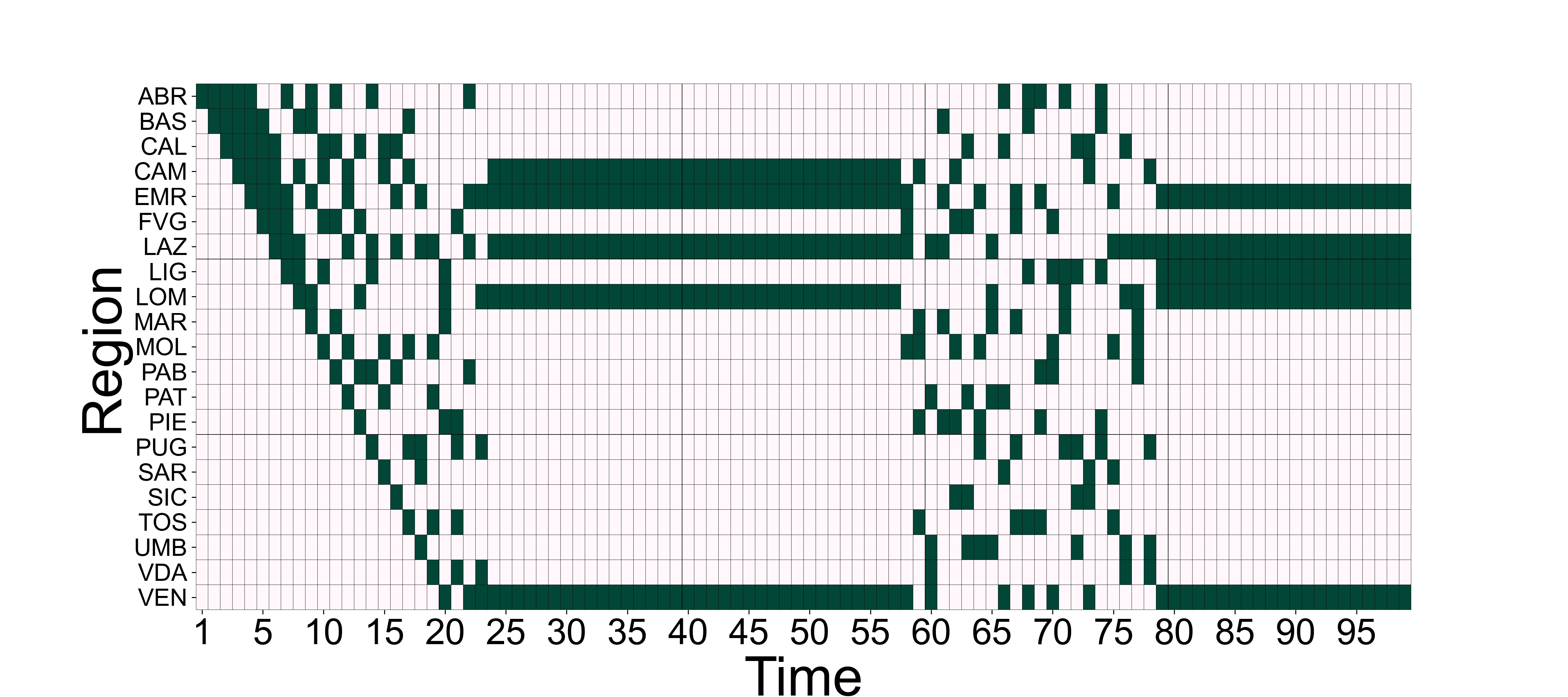}
}
\caption{Selected regions on each day of the experiment.}
\label{fig:chosen-regions}
\end{figure}

In this section, we provide the results of applying our algorithm, to the Covid-19 outbreak dataset of daily new infected cases during the pandemic in different regions within Italy.\footnote{https://github.com/pcm-dpc/COVID-19} The goal is to find a subset of regions with the highest contribution to the spread of the virus in the country in a non-stationary period. We use the \textit{overall reward} ${y}_{i}$ for the \textit{overall daily new cases} in region $i$. Besides, we use the \textit{instantaneous reward} ${b}_{i}$ for the \textit{region-specific daily new cases} in region $i$. The data of the period from $3^{rd}$ July $2020$, to $10^{th}$ October $2020$ was used. We pre-process the dataset following \cite{nourani2022linear}; nevertheless, we use a $14$-day moving average instead of a $7$-day moving average. Instead of the $L^{1}\text{-norm}$ in (\ref{eq:Optimization-A}), we use the Directed Total Variation (DTV) $ \sum_{i,j \in \mathcal{K}} \mathbf{W}[i,j] \sum_{h= 1, \dots, t} \left [ \mathbf{Y}[i,h] - \mathbf{Y}[j,h] \right]^{+}$ regularizer \cite{sardellitti2017graph}, where $ \left[ y \right]^{+} = \text{max} \left\{ y, 0 \right \}$. Since the causal spread of the disease might create cycles, we allow cyclic graphs as the result of the optimization problem (\ref{eq:Optimization-A}). Considering that the ground truth graphs are not available, we use a cross-validation technique to tune the regularization parameter $\lambda_{1}$. We split the data into $10$ subsets of $10$ consecutive days. In each subset, one day is chosen uniformly at random to be included in the validation set, while the remaining $9$ days are added to the train set. We calculate the prediction error at each time $t$ by $\textit{Error}(t) = \frac{1}{K | \mathfrak{v}(t)|} \sum_{i \in \mathfrak{v}(t)} \left\| \mathbf{y}_{i} - \hat{\mathbf{y}}_{i} \right\|_{1}$
%
where $\mathfrak{v}(t)$ is the validation set at time $t$ with cardinality $|\mathfrak{v}(t)|$. Besides, $\mathbf{y}_{i}$ and $\hat{\mathbf{y}}_{i}$ are respectively the validation data, and the corresponding predicted value using the estimated graph for day $i$. \Cref{fig:sig-reconstruct} compares the ground truth overall daily new cases and the predicted total daily new cases using the estimated graph in $3$ days of the Covid-19 outbreak in our validation data.\footnote{Due to space limitations, we use abbreviations for region names. Table $1$ in supplementary material lists the abbreviations together with the original names of the regions.}~According to \Cref{fig:sig-reconstruct}, our algorithm estimates the data for each region efficiently. This helps the agent to find the optimal decision vector. Regarding that the benchmarks need the prior knowledge of the causal graph, this real-world application highlights the drawbacks of the benchmarks. Considering the impacts of geographical factors on Covid-19 cases \cite{wang2022impact}, we divide the country into $4$ clusters, using \textit{graph-based clustering} library of \textit{Python}, based on Euclidean distances between regional capitals. In \Cref{fig:chosen-regions}, we show the regions that PS-SEM-UCB-Gr selects over time. On each day, the selected regions are highlighted by dark rectangles. PS-SEM-UCB-Gr finds changes in the distribution of the region-specific daily new cases of different regions belonging to each group. Consequently, it restarts the UCB procedure for all the groups within the period $t=58$ and $t=79$. Due to space limitations, the details about the groupings and their change-detection times are mentioned in the supplementary. We see that selected subsets of regions before and after the restart of the algorithm are different due to newly calculated UCB indices after the restarts. This shows how the main contributors to the spread of the virus changed from one stationary segment to the next. 
%
%
\section{Conclusion}
\label{sec:conclusion}
In this paper, we developed a piecewise stationary combinatorial semi-bandit framework with causally related rewards. We developed a decision-making policy that follows distribution- and causal graph changes to adapt the decisions. We introduced a new alternative for the restarting process of bandit algorithms in structured environments under piecewise stationary settings. We proved that PS-SEM-UCB-Gr achieves a sublinear regret bound. The experiments showed the superior performance of PS-SEM-UCB-Gr compared to several state-of-the-art combinatorial algorithms. Our regret analysis clarifies the effects of global and local restarts as special cases of group restarts. It clarifies the importance of using relationships amongst base arms' distributions for the purpose of grouping of arms to minimize the regret incurred by the restarting process in group restarts. As for future research direction, we aim at studying our problem under the presence of noise in the SEM.
\section*{Acknowledgements}This work was partially funded by the Deutsche Forschungsgemeinschaft (DFG, German Research Foundation) under Germany’s Excellence Strategy – EXC number 2064/1 – Project number 390727645, and by Grant 16KISK035 from the German Federal Ministry of Education and Research (BMBF). We are grateful to Sofien Dhouib for fruitful discussions and comments.
\bibliography{ecai}
\end{document}